\documentclass[11pt,a4paper]{article}
\let\maketitle\relax
\PassOptionsToPackage{hyphens}{url}\usepackage[breaklinks=true]{hyperref}
\usepackage{acl2021}
\usepackage{times}
\usepackage{latexsym}
\usepackage[hyphenbreaks]{breakurl}

\usepackage{microtype}

\aclfinalcopy 

\newcommand*{\affaddr}[1]{#1} 
\newcommand*{\affmark}[1][*]{\textsuperscript{#1}}
\newcommand*{\email}[1]{\texttt{#1}}

\title{ViNMT: Neural Machine Translation Tookit}

\author{
Nguyen Hoang Quan\affmark[1]\affmark[*], Nguyen Thanh Dat\affmark[1]\affmark[*], Nguyen Hoang Minh Cong\affmark[1]\affmark[*], Nguyen Van Vinh\affmark[1]\affmark[†], \\\AND Ngo Thi Vinh\affmark[2], Nguyen Phuong Thai\affmark[1], and Tran Hong Viet\affmark[3]\\
\affaddr{\affmark[1]University of Engineering and Technology - Hanoi VNU}\\
\email{\{quan94fm,17020049,congnhm,vinhnv,thainp\}@vnu.edu.vn}\\
\affaddr{\affmark[2]University of Information and Communication Technology - Thai Nguyen University}\\
\email{ntvinh@ictu.edu.vn}\\
\affaddr{\affmark[3]University of Economics-Technical For Industries}\\
\email{thviet@uneti.edu.vn} \\ \\
}

\date{Hanoi, 29/07/2022}

\begin{document}

\maketitle{}
{\let\thefootnote\relax\footnote{{* Authors with equal contributions}}
\footnote{{† Corresponding Author}}}
\begin{abstract}
We present an open-source toolkit for
neural machine translation (NMT). The new toolkit is mainly based on the vaulted Transformer \citep{vaswani2017attention} along with many other improvements detailed below, in order to create a self-contained, simple to use, consistent and comprehensive framework for Machine Translation tasks of various domains. It is tooled to support both bilingual and multilingual translation tasks, starting from building the model from respective corpora, to inferring new predictions or packaging the model to serving-capable JIT format. The source code and data are available at \url{https://github.com/KCDichDaNgu/MultilingualMT-UET-KC4.0}.
\end{abstract}

\section{Introduction}
With the emergence of neural-network based Machine Learning models and their wide application, Natural Language Processing tasks in general and Machine Translation (MT) task in particular had benefitted from this new architecture, receiving massive gains in both translation quality and fluency.

In particular, the current state-of-the-art architecture for MT task is Transformer \citep{vaswani2017attention} which features various extensions of the Attention concept adapted to older neural network models \citep{bahdanau2014neural}, \citep{luong-etal-2015}. Variants of the Transformer architecture had been investigated, such as changes in its positional encoding \citep{T5}, changes to its attention to accommodate very long sequence \citep{omninet}, or even adaptations that only employ the autoregressive decoder \citep{gpt3}, but so far the base Transformer model is robust, competitive and straightforward for the vast majority of translation cases. 

In a similar vein, currently production-level translation models often leverage multiple translation corpora to enable greater proficiency, achieving increasingly better translation compared to bilingual models \citep{massive-multilingual} and enable even Zero-Shot training for rare language pairs that had no prior training data \citep{google-multilingual}. Most known Transformer frameworks available do not explicitly support this, and thus we designed our engine to support this goal to the best of our ability.

Amongst open-source engines capable of supporting Machine Translation tasks, there are OpenNMT \cite{klein-etal-2020-opennmt} which boasts a well defined structure and have different versions supporting different underlying neural frameworks; Marian \cite{mariannmt} which is written in C++ for absolute performance and minimal dependency; and Fairseq \cite{ott-etal-2019-fairseq} which provide many state-of-the-art models and improvements. However, in OpenNMT and Fairseq's case, the breadth of their works makes it hard for new users to experiment on modifying the structure, while Marian's compactness is offset by the lack of readability in C++. 

Our engine is designed with two qualities in mind: \textbf{Modularity} and \textbf{Coherency}. First, excluding running scripts and unimportant support functions, the rest of our engine is designed to be built revolving around replaceable modules with known interface, thus enabling simple inheritance, replacement or upgrading for parts of the neural network models. Second, modules are strictly designed for one purpose and one purpose only, thus prevent unnecessary bloating in a single class and the consequent pain of modifying one. 

\section{Technical Details}
\subsection{Neural Machine Translation}
Neural Machine Translation is a fully-automated translation based on neural network. \citep{2014arXiv1406.1078C} suggest a new architectural and called Sequence-to-Sequence (Seq2Seq) model. They apply memory unit such as Long Short-Term Memory (LSTM) or Gated Recurrent Units (GRU) to surmount the exploding or vanishing gradient problem in recurrent networks. The architecture includes an encoder and decoder. The one Encoder uses to present the sentence in the source language with n tokens X = ($x_1$; $x_2$; ...; $x_n$) and a decoder to generate the predicted sentence in the target language with m tokens Y = ($y_1$; $y_2$; ...; $y_m$). The model evaluates the conditional probability $P(y_j \mid Y_{<j}, X)$ generating output sentence Y when the input sequence X is given:

\[ P(Y \mid X) = \prod_{j=1}^{m+1} P(y_j \mid Y_{<j}, X)\]

In the Seq2seq model, the global attentions mechanism \citep{luong-etal-2015} is considered to compute alignment attention from source sentences to corresponding target sentences.
However, due to the Seq2seq model computing the probability sequentially, they have limited parallelization in the training process. 

\subsection{Transformer-based NMT}

To solve the parallelization problem, the transformer architecture for machine translation
is mentioned for the first time by \citep{vaswani2017attention}. They are noted to be highly parallelizable as well as better in translating long sentences. Instead of using GRU or LSTM units to encode source sentences sequentially, they use encoder layers that help the encoder look at other words in the input sentence as it encodes a specific word. This architecture allows to train much faster and gives a better quality compared to RNN architecture. The self-attention mechanism as the
following:

\[ \emph{Attention(Q, K, V)} = \emph{Softmax}(\frac{QK^T}{\sqrt{d}}) V\]

Here, K (key), Q (query), V (value) is the outputs of the encoder or decoder to present tokens in the input sentence, and d is the size of the input.
And then, the Transformer is trained to optimize parameters $\theta$ by minimizing the maximum likelihood of the sentences pairs: 

\[ \emph{L($\theta$)} = \frac{1}{T} \sum_{k=1}^{T} log P(Y \mid X, \theta) \]

where T is the number of sentence pairs in the multilingual corpus. We use transformer-based for all our experiments. 

\subsection{Multilingual Neural Machine System}
Multilingual neural machine translation (MNMT) system can translate many language pairs, which has been useful in improving translation quality as a result of translation knowledge transfer (transfer learning), even low-resource, zeros-shot issues. (MNMT) has some strategy to translate between many languages: 
\begin{itemize}
    \item \textbf{(1)} Many to many \citep{ha2016multilingual}: from many sources to many target languages;
    \item \textbf{(2)} Many to one \citep{gu2019improved}: from many sources to a target language;
    \item \textbf{(3)} One to many \citep{wang2018three}: from a sources to many target languages;
\end{itemize}
Our motivation is to improve low-resource translation tasks and focus on translating from many languages to one language, so we systems are the same as the case (2). In a MNMT system from many to one, the objective function uses maximum likelihood estimation on the whole parallel pairs:

\[ \emph{L($\theta$)} = \frac{1}{K} \sum_{m=1}^{M} \sum_{k=1}^{K} log P(Y^{(m, k)} \mid X^{(m, k)}, \theta) \]
Where M is the number of languages and K is the total number of sentences in languages m. Of course, the vocabulary of the source side is mixed from all source language: 
\[ \emph{V} = \sum_{m=1}^{M} V_m \]

\textbf{Code-mixed Language:} The benefit of shared vocab in MNMT has been shown by \cite{gu2019improved}  that if the languages shared the same alphabet and had many similar words, MNMT system can get many advantages of common features of languages. In order to do that they may share the same subwords in all language pairs. This significantly reduces the number of rare words in the MT systems. In our experiments, we choose English, Chinese, Khmer, and Lao are source languages and hope that they can share many tokens with each other to translate into Vietnamese.

\section{Implementation}

The first section details the overall structure of our engine to adhere to our outlined qualities specified above. In general, most components are built as interlinked modules that can be replaced or modified on its parent modules with minimal effort using python's optionality.

The accompanying sections describe various techniques we used to improve the efficiency and accuracy of our translation system in both the training and inference phases. Besides, we also build a module that allows packaging the model to simplify the serving process. We heavily rely on PyTorch framework to implement all these techniques.

\subsection{System Design}
In order to support the technical system above, our code base is designed from an Object-Oriented point of view, with interchangeable base classes following PyTorch's \textbf{nn.Module} class. This modular approach allows easy and simple modification when experimenting or upgrading specific parts of the system, since in most cases we can simply inherit the base class that needed to be changed and use our new class as an initiation argument.

Our system revolves around three main important classes:
\begin{itemize}
 \item the \textbf{Layer} class, which are layers that directly involve in neural network's training and processing. As such, they are exclusively contained by the neural network modules in the system such as the \textbf{Encoder} and \textbf{Decoder} modules. 
 \item the \textbf{Module} class, which make up the bulk of our coding base. A Module represents a specific duty to be filled, for example, our Transformer base model currently contain a \textbf{Loader} object for converting text data to numeric form and vice versa; a pair of \textbf{Encoder-Decoder} objects for the main translation process, and an additional \textbf{DecoderStrategy} object to support the model during autoregressive inference.
 \item the \textbf{Model} class, which contains all the respective modules required when training and inferring. This packaging allows (1) the highest level of user interference into translating process without the need of understanding supporting codes such as the \textbackslash{bin} or \textbackslash{utils} code; and (2) simple usage during \textbf{TorchScript} serving, as the converted object would be a stand-alone module that doesn't have to worry about re-implementing specific non-neural modules.
\end{itemize} 

Using an example to demonstrate the modularity of our system, assuming that we have a new \textit{Decoder} object to put into our original model, all we have to do is to replace the corresponding \textit{decoder{\_}cls} argument within the constructor function of the \textit{Model} class. This can be easily accomplished by writing a lambda function in the model name list specified within  \textbackslash{models} \textbackslash{}{\_}{\_}init{\_}{\_}.py

In addition, our code base contains several additional code and resources: \textbackslash{bin} contains command line scripts to run the system, \textbackslash{config} contains basic configuration files which contain model hyperparameters that will be preserved during runs, \textbackslash{web} contains serving scripts and resources, \textbackslash{utils} contains extra support functions that don't fit other folders. 

\subsection{Training}

\paragraph{Training paradigm}
Our system allows optionally training models by epochs or steps:
\begin{itemize}
    \item If \textbf{epoch mode} is chosen, the model is trained in a specified number of epochs. For each epoch, the training data is shuffled and partitioned into several batches based on which batching algorithm is used. After that, these batches are fed into the model for training.
    
    \item If \textbf{sampling mode} is chosen, the model is trained in a specified number of steps. For each step, several training examples (sentence pairs) are selected based on a specified distribution to form a training batch. This mode is more suitable when we want to train a multilingual model while there is a significant different training data size between language pairs.
\end{itemize}

\paragraph{Automatic Mixed Precision}
The Pytorch framework also provides methods for mixed precision. When this method is applied, some operations use float (32-bit) datatype and other operaions use half-float (16-bit) datatype. Some operations, like linear layers, are much faster in float 16-bit. Other operations, like reductions, often require the dynamic range of float 32-bit. Mixed precision tries to match each operation to its appropriate datatype, which can reduce our network’s runtime and memory usage.

\subsection{Inference}

We use the standard beam search algorithm to generate texts on the target side. The baseline configuration includes beam size \(k = 4\), length penalty \(alpha = 0.6\) and maximum number of decode steps is 128. There are some techniques we can use to speed up the inference process and reduce memory usage:

\paragraph{Attention caching}
Transformer allows us to train much faster than RNN because inputs are processed in parallel. However, this benefit does not work in the inference phase because the model has to sequentially generate each token in each decoding step. Thus, the Transformer model needs entire target tokens generated from the previous decoding steps as a decoder input to compute the next target token. Therefore, all \textbf{Query} (Q), \textbf{Key} (K), \textbf{Value} (V) vectors from previous steps are re-calculated.

Instead of repeatedly compute Q, K, V in every decoding step, we want to store them in the memory so that they can be used for further calculations. This idea works because all Q, K, V vectors derived from the previous target tokens do not change after each decoding step. Caching these Q, K, V vectors makes calculating \textbf{Self-Attention} and \textbf{Cross-Attention} matrix in a different way. In details:

\begin{itemize}
    \item To calculate Self-Attention matrix, we need concatenate all the previous K, V vectors to form a complete K, V tensor and perform matrix multiplication with the current Q vector. The result is an attention tensor that attends from the current target token to previous target tokens. This calculation works perfectly when we apply causal masking for self-attention tensor on decoder side.
    
    \item To calculate Cross-Attention matrix, we simply perform matrix multiplication between the current Q vector and K, V tensors from encoder output. The result is an attention tensor that attends from current target token to all source tokens.
\end{itemize}

\paragraph{Sentences sorting}
Sorting source sentences by their length (calculated by the number of tokens in these sentences) and dynamically batching such as the total number of tokens does not exceed 4069s. By applying this technique, the model uses less memory compared to the regular batching algorithm (batching by the number of sentences) because we do not have to add too many pad tokens to every sentence in the batch.

\subsection{Serving}
We want the deployment and training process are independent as much as possible, the models can be deployed on any environment besides Python. Therefore, we use \textbf{TorchScript} which is a way to create serializable and optimize models from PyTorch code. Any TorchScript program can be saved from a Python process and loaded in a process where there is no Python dependency. It transitions a model from a pure Python program to a TorchScript program that can be run independently from Python and then export the model via TorchScript to a production environment where Python programs may be disadvantageous for performance and multi-threading reasons.

\section{Experiments}

This section describes our experiments using our own data.

\subsection{Dataset}
We crawled data from the news domain for four language pairs English-Vietnamese, Chinese-Vietnamese, Laos-Vietnamese and Khmer-Vietnamese. The details of those datasets are described in Table 1.

\begin{table}[ht]
\caption{The bililingual dataset in our experiments} 
\centering 
\begin{tabular}{c c c c} 
\hline\hline 
Language pairs & Train & Valid & Test \\ [0.5ex] 
\hline 
En - Vi & 3M & 1553 & 1268 \\ 
Cn - Vi & 675K & 1000 & 1000 \\
Lo - Vi & 83K & 1000 & 1000 \\
Km - Vi & 70K & 1000 & 1000 \\
\hline 
\end{tabular}
\label{table:nonlin} 
\end{table}
\subsection{Preprocessing}

All parallel texts were tokenized and truncated using sentencepiece scripts, and then they are applied to Sennrich’s BPE (Sennrich et al., 2016). We explore 32K operators are learned to generate BPE codes for all languages. 

For Vietnamese, we only apply Moses’s scripts for tokenization and true-casing.

\subsection{Systems and Training}

We implement our NMT systems from zeros-base to train all our experiments. The same settings are used
for all experiments. 

We trained our Transformer model using the number of encoder 12, decoder layers are 6, 8 head is used, \(d_{model}\) is 512, dropout value is 0.1, batch size of 64, learning rate value is 0.4 with the aid of Adam optimizer. The learning rate has warmup updates by 8000 steps and label smoothing value is 0.1.

We evaluate the quality of two systems (1) bilingual system, (2) multilingual system.
\section*{(1) Bililingual system}
We train systems based on separate bilingual data of each language pair. We explore the best model to decode the test data for comparison purposes in our tests. The English - Vietnamese model we trained for 30 epochs, Chinese - Vietnamese, Lao - Vietnamese, Khmer - Vietnamese for 20 epochs. We train bilingual system for the baseline and compare with multilingual system.
\section*{(2) Multilingual system}
We concatenate training sets for all language pairs in order to construct the new sets: English, Chinese, Khmer, Lao $\,\to\,$Vietnamese. We train the system using those data for the same number of epochs. And then, we compare with bilingual translation results and see that improvement of +4.06 BLEU points
on English → Vietnamese translation task, another one of +0.56 BLEU points on Chinese → Vietnamese, +4.19 BLEU points on Lao → Vietnamese, and +3.18 BLEU points on Khmer → Vietnamese.

\subsection{Results}

The results of our experiments are shown in table 2 and table 3.
\begin{table}[ht]
\caption{BLEU scores for bililingual system} 
\centering 
\begin{tabular}{c c c c} 
\hline\hline 
Language pairs & BLEU \\ [0.5ex] 
\hline 
En - Vi & 31.77 \\ 
Cn - Vi & 27.96 \\
Lo - Vi & 16.29 \\
Km - Vi & 20.78 \\
\hline 
\end{tabular}
\label{table:nonlin} 
\end{table}

\begin{table}[ht]
\caption{BLEU scores for multilingual system} 
\centering 
\begin{tabular}{c c c c} 
\hline\hline 
Language pairs & BLEU \\ [0.5ex] 
\hline 
En - Vi & 34.98 \\ 
Cn - Vi & 28.62 \\
Lo - Vi & 18.94 \\
Km - Vi & 23.44 \\

\hline 
\end{tabular}
\label{table:nonlin} 
\end{table}

\section{Conclusion}

We have built a research toolkit for NMT that design efficiency and modularity. And release all code for community NLP, specially, machine translation of Vietnam. We find that the large corpus bilingual can furthermore enhance a multilingual NMT system. The MNMT system significantly reduces the number of rare words in the MT systems. Nevertheless, the rare word issue is still a challenge in NMT.

In the future, we will continue develop ViNMT to achieve strong MT results at up-to-date research.

\section*{Acknowledgments}

This work has been supported by Ministry of Science and Technology of Vietnam under Program KC 4.0, No. KC-4.0.12/19-25.

\bibliographystyle{acl_natbib}
\bibliography{anthology,acl2021}

\begin{thebibliography}{15}
\expandafter\ifx\csname natexlab\endcsname\relax\def\natexlab#1{#1}\fi

\bibitem[{Aharoni et~al.(2019)Aharoni, Johnson, and
  Firat}]{massive-multilingual}
Roee Aharoni, Melvin Johnson, and Orhan Firat. 2019.
\newblock \href {http://arxiv.org/abs/1903.00089} {Massively multilingual
  neural machine translation}.
\newblock \emph{CoRR}, abs/1903.00089.

\bibitem[{Bahdanau et~al.(2015)Bahdanau, Cho, and Bengio}]{bahdanau2014neural}
Dzmitry Bahdanau, Kyunghyun Cho, and Yoshua Bengio. 2015.
\newblock Neural machine translation by jointly learning to align and
  translate.
\newblock In \emph{3rd International Conference on Learning Representations,
  {ICLR} 2015, San Diego, CA, USA, May 7-9, 2015, Conference Track
  Proceedings}.

\bibitem[{Brown et~al.(2020)Brown, Mann, Ryder, Subbiah, Kaplan, Dhariwal,
  Neelakantan, Shyam, Sastry, Askell, Agarwal, Herbert{-}Voss, Krueger,
  Henighan, Child, Ramesh, Ziegler, Wu, Winter, Hesse, Chen, Sigler, Litwin,
  Gray, Chess, Clark, Berner, McCandlish, Radford, Sutskever, and
  Amodei}]{gpt3}
Tom~B. Brown, Benjamin Mann, Nick Ryder, Melanie Subbiah, Jared Kaplan,
  Prafulla Dhariwal, Arvind Neelakantan, Pranav Shyam, Girish Sastry, Amanda
  Askell, Sandhini Agarwal, Ariel Herbert{-}Voss, Gretchen Krueger, Tom
  Henighan, Rewon Child, Aditya Ramesh, Daniel~M. Ziegler, Jeffrey Wu, Clemens
  Winter, Christopher Hesse, Mark Chen, Eric Sigler, Mateusz Litwin, Scott
  Gray, Benjamin Chess, Jack Clark, Christopher Berner, Sam McCandlish, Alec
  Radford, Ilya Sutskever, and Dario Amodei. 2020.
\newblock \href {http://arxiv.org/abs/2005.14165} {Language models are few-shot
  learners}.
\newblock \emph{CoRR}, abs/2005.14165.

\bibitem[{{Cho} et~al.(2014){Cho}, {van Merrienboer}, {Gulcehre}, {Bahdanau},
  {Bougares}, {Schwenk}, and {Bengio}}]{2014arXiv1406.1078C}
Kyunghyun {Cho}, Bart {van Merrienboer}, Caglar {Gulcehre}, Dzmitry {Bahdanau},
  Fethi {Bougares}, Holger {Schwenk}, and Yoshua {Bengio}. 2014.
\newblock \href {http://arxiv.org/abs/1406.1078} {{Learning Phrase
  Representations using RNN Encoder-Decoder for Statistical Machine
  Translation}}.
\newblock \emph{arXiv e-prints}, page arXiv:1406.1078.

\bibitem[{Gu et~al.(2019)Gu, Wang, Cho, and Li}]{gu2019improved}
Jiatao Gu, Yong Wang, Kyunghyun Cho, and Victor O.~K. Li. 2019.
\newblock \href {http://arxiv.org/abs/1906.01181} {Improved zero-shot neural
  machine translation via ignoring spurious correlations}.

\bibitem[{Ha et~al.(2016)Ha, Niehues, and Waibel}]{ha2016multilingual}
Thanh-Le Ha, Jan Niehues, and Alexander Waibel. 2016.
\newblock \href {http://arxiv.org/abs/1611.04798} {Toward multilingual neural
  machine translation with universal encoder and decoder}.

\bibitem[{Johnson et~al.(2016)Johnson, Schuster, Le, Krikun, Wu, Chen, Thorat,
  Vi{\'{e}}gas, Wattenberg, Corrado, Hughes, and Dean}]{google-multilingual}
Melvin Johnson, Mike Schuster, Quoc~V. Le, Maxim Krikun, Yonghui Wu, Zhifeng
  Chen, Nikhil Thorat, Fernanda~B. Vi{\'{e}}gas, Martin Wattenberg, Greg
  Corrado, Macduff Hughes, and Jeffrey Dean. 2016.
\newblock \href {http://arxiv.org/abs/1611.04558} {Google's multilingual neural
  machine translation system: Enabling zero-shot translation}.
\newblock \emph{CoRR}, abs/1611.04558.

\bibitem[{Junczys-Dowmunt et~al.(2018)Junczys-Dowmunt, Grundkiewicz, Dwojak,
  Hoang, Heafield, Neckermann, Seide, Germann, Fikri~Aji, Bogoychev, Martins,
  and Birch}]{mariannmt}
Marcin Junczys-Dowmunt, Roman Grundkiewicz, Tomasz Dwojak, Hieu Hoang, Kenneth
  Heafield, Tom Neckermann, Frank Seide, Ulrich Germann, Alham Fikri~Aji,
  Nikolay Bogoychev, Andr\'{e} F.~T. Martins, and Alexandra Birch. 2018.
\newblock \href {http://www.aclweb.org/anthology/P18-4020} {Marian: Fast neural
  machine translation in {C++}}.
\newblock In \emph{Proceedings of ACL 2018, System Demonstrations}, pages
  116--121, Melbourne, Australia. Association for Computational Linguistics.

\bibitem[{Klein et~al.(2020)Klein, Hernandez, Nguyen, and
  Senellart}]{klein-etal-2020-opennmt}
Guillaume Klein, Fran{\c{c}}ois Hernandez, Vincent Nguyen, and Jean Senellart.
  2020.
\newblock \href {https://www.aclweb.org/anthology/2020.amta-research.9} {The
  {O}pen{NMT} neural machine translation toolkit: 2020 edition}.
\newblock In \emph{Proceedings of the 14th Conference of the Association for
  Machine Translation in the Americas (Volume 1: Research Track)}, pages
  102--109, Virtual. Association for Machine Translation in the Americas.

\bibitem[{Luong et~al.(2015)Luong, Pham, and Manning}]{luong-etal-2015}
Thang Luong, Hieu Pham, and Christopher~D. Manning. 2015.
\newblock \href {https://doi.org/10.18653/v1/D15-1166} {Effective approaches to
  attention-based neural machine translation}.
\newblock pages 1412--1421.

\bibitem[{Ott et~al.(2019)Ott, Edunov, Baevski, Fan, Gross, Ng, Grangier, and
  Auli}]{ott-etal-2019-fairseq}
Myle Ott, Sergey Edunov, Alexei Baevski, Angela Fan, Sam Gross, Nathan Ng,
  David Grangier, and Michael Auli. 2019.
\newblock \href {https://doi.org/10.18653/v1/N19-4009} {fairseq: A fast,
  extensible toolkit for sequence modeling}.
\newblock In \emph{Proceedings of the 2019 Conference of the North {A}merican
  Chapter of the Association for Computational Linguistics (Demonstrations)},
  pages 48--53, Minneapolis, Minnesota. Association for Computational
  Linguistics.

\bibitem[{Raffel et~al.(2019)Raffel, Shazeer, Roberts, Lee, Narang, Matena,
  Zhou, Li, and Liu}]{T5}
Colin Raffel, Noam Shazeer, Adam Roberts, Katherine Lee, Sharan Narang, Michael
  Matena, Yanqi Zhou, Wei Li, and Peter~J. Liu. 2019.
\newblock \href {http://arxiv.org/abs/1910.10683} {Exploring the limits of
  transfer learning with a unified text-to-text transformer}.
\newblock \emph{CoRR}, abs/1910.10683.

\bibitem[{Tay et~al.(2021)Tay, Dehghani, Aribandi, Gupta, Pham, Qin, Bahri,
  Juan, and Metzler}]{omninet}
Yi~Tay, Mostafa Dehghani, Vamsi Aribandi, Jai~Prakash Gupta, Philip Pham, Zhen
  Qin, Dara Bahri, Da{-}Cheng Juan, and Donald Metzler. 2021.
\newblock \href {http://arxiv.org/abs/2103.01075} {Omninet: Omnidirectional
  representations from transformers}.
\newblock \emph{CoRR}, abs/2103.01075.

\bibitem[{Vaswani et~al.(2017)Vaswani, Shazeer, Parmar, Uszkoreit, Jones,
  Gomez, Kaiser, and Polosukhin}]{vaswani2017attention}
Ashish Vaswani, Noam Shazeer, Niki Parmar, Jakob Uszkoreit, Llion Jones,
  Aidan~N. Gomez, Lukasz Kaiser, and Illia Polosukhin. 2017.
\newblock \href {http://arxiv.org/abs/1706.03762} {Attention is all you need}.

\bibitem[{Wang et~al.(2018)Wang, Zhang, Zhai, Xu, and Zong}]{wang2018three}
Yining Wang, Jiajun Zhang, Feifei Zhai, Jingfang Xu, and Chengqing Zong. 2018.
\newblock \href {https://doi.org/10.18653/v1/D18-1326} {Three strategies to
  improve one-to-many multilingual translation}.
\newblock In \emph{Proceedings of the 2018 Conference on Empirical Methods in
  Natural Language Processing}, Brussels, Belgium. Association for
  Computational Linguistics.

\end{thebibliography}


\end{document}